\documentclass[10pt,twocolumn,letterpaper]{article}

\usepackage{cvpr}
\usepackage{multirow}
\usepackage{caption}
\usepackage{indentfirst}
\usepackage[T1]{fontenc}

\usepackage{etoolbox}
\makeatletter
\patchcmd{\abstract}{\noindent}{}{}{}
\makeatother
\definecolor{cvprblue}{rgb}{0.21,0.49,0.74}
\usepackage[pagebackref,breaklinks,colorlinks,allcolors=cvprblue]{hyperref}

\title{SpaceMind: Camera-Guided Modality Fusion for Spatial Reasoning in Vision-Language Models}

\author{
\parbox{\textwidth}{\centering
Ruosen Zhao$^{1,2,*}$ \quad
Zhikang Zhang$^{1,*}$ \quad
Jialei Xu$^{1}$ \quad
Jiahao Chang$^{2}$ \quad
Dong Chen$^{1,3}$ \quad
Lingyun Li$^{1}$ \quad
Weijian Sun$^{1}$ \quad
Zizhuang Wei$^{1,\dagger}$\\[0.35em]
$^{1}$ Huawei \quad
$^{2}$ The Chinese University of Hong Kong, Shenzhen \quad
$^{3}$ The University of Hong Kong\\[0.25em]
{\tt\small
\{ruosenzhao, jiahaochang\}@link.cuhk.edu.cn, \\
olichen@connect.hku.hk, \\
\{zhangzhikang7, xujialei5, lilingyun.hw, sunweijian, weizizhuang\}@huawei.com
}\\[0.2em]
{\footnotesize $^\ast$Equal contribution.\quad $^\dagger$Corresponding author.}
}
}

\begin{document}
\maketitle

\begin{abstract}
Large vision-language models (VLMs) show strong multimodal understanding but still struggle with 3D spatial reasoning, such as distance estimation, size comparison, and cross-view consistency. Existing 3D-aware methods either depend on auxiliary 3D information or enhance RGB-only VLMs with geometry encoders through shallow feature fusion.
We propose SpaceMind, a multimodal large language model explicitly designed for spatial reasoning solely from RGB inputs. The model adopts a dual-encoder architecture, integrating VGGT as a spatial understanding encoder and InternViT as a 2D visual encoder.
The key idea is to treat the camera representation as an active guiding modality rather than passive metadata. Specifically, SpaceMind introduces a lightweight Camera-Guided Modality Fusion module before the language model to replace shallow fusion. It applies camera-conditioned biasing to spatial tokens, assigns query-independent weights reflecting their geometric importance, and uses the camera embedding to gate the fused representation.
Empirically, SpaceMind establishes new \textbf{state-of-the-art} results on \textbf{VSI-Bench}, \textbf{SQA3D} and \textbf{SPBench}, surpassing both open and proprietary systems on VSI-Bench and SPBench by large margins and achieving state-of-the-art performance on SQA3D.
These results demonstrate that camera-guided modality fusion is an effective and practical inductive bias for equipping VLMs with genuinely spatially grounded intelligence.
We will release code and model checkpoints to support future research.

\begin{figure}[t]
    \centering
    \includegraphics[width=\columnwidth]{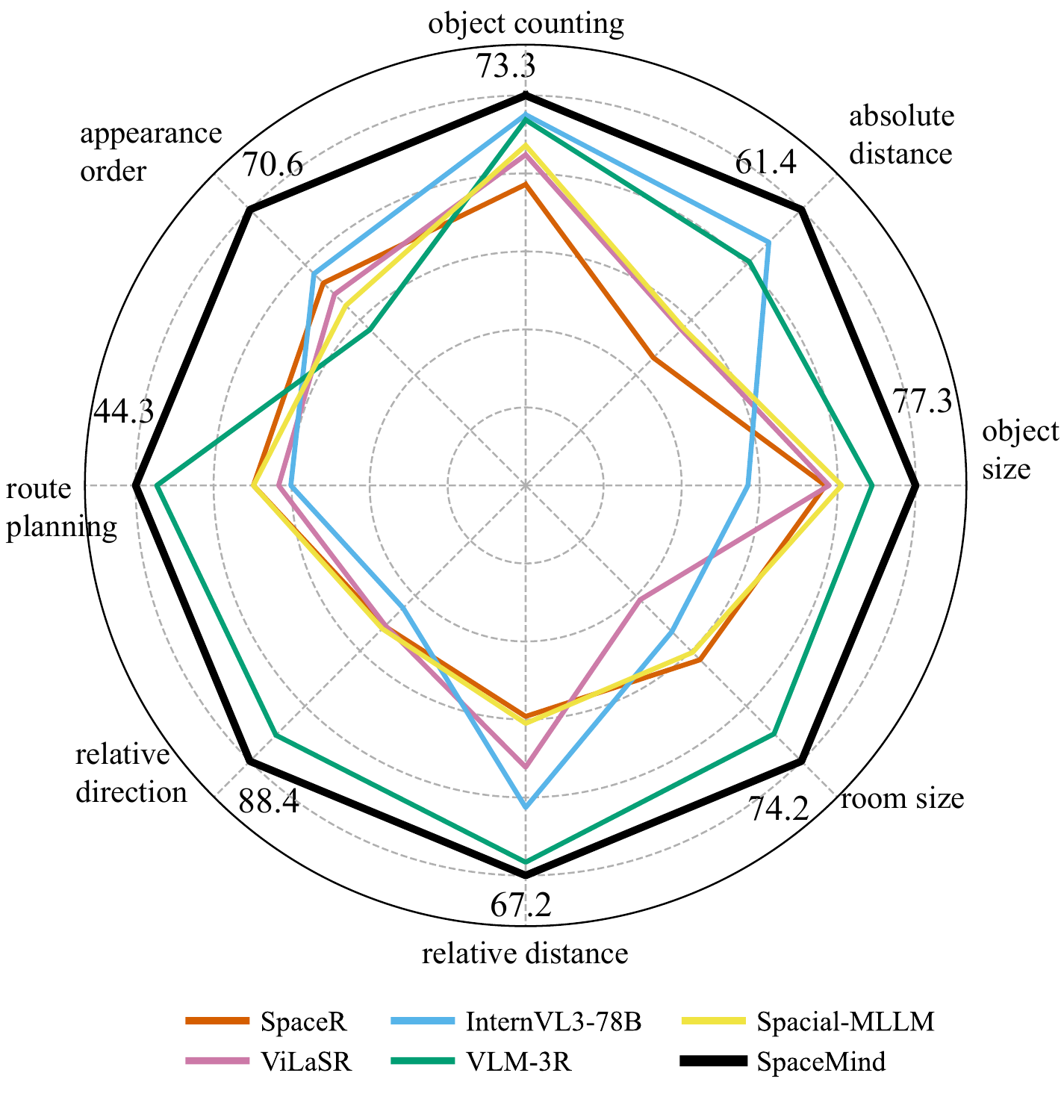}
    \caption{Performance on VSI-Bench across different spatial reasoning categories. SpaceMind achieves consistently strong visuospatial intelligence compared to existing systems.}
    \label{fig:radar}
\end{figure}

\end{abstract}

\section{Introduction}

Humans perceive space not merely by seeing, but by understanding \emph{from where} they see.
This coupling between visual observation and viewpoint underlies our ability to estimate distances, compare sizes, infer connectivity, and navigate unfamiliar environments.
Endowing multimodal models with such \textbf{spatial intelligence} \cite{yang2025vsi} is crucial for a wide range of applications.
However, despite remarkable progress in large language models (LLMs) \cite{naveed2023comprehensive,wei2022emergent,begus2023large,zhang2024unveiling,kassner2023language,radford2018improving,radford2019language,brown2020language,touvron2023llama,touvron2023llama2} 
and multimodal large language models (MLLMs) \cite{bai2023qwen,gemini2023gemini,hurst2024gpt4o,gemini2024gemini15,liu2024visual,alayrac2022flamingo,li2023blip2,bai2023qwenvl,chen2024internvl} on 
open-ended reasoning, instruction following, and long-context understanding, 
recent studies consistently show that they struggle with explicitly 3D-aware tasks, such as metric comparison, layout inference, and multi-view consistency.

Existing approaches to spatially grounded multimodal reasoning can be roughly grouped into two families.
The first family augments language or vision-language models with \emph{explicit 3D inputs}, such as point clouds, depth maps, reconstructed meshes, or BEV/voxel features
\cite{zhu2024pq3d,zhu2023vista,yuan2025scener1,ma2024leo,man2024sig3d,wang2025scenellm,deng20253dllava,huang2024chatscene,zhu2025llava3d,zheng2025video3dllm}.
These methods leverage depth sensors or multi-view reconstruction pipelines to build 3D scene representations, which are then encoded and aligned with language.
They can capture rich geometric details, but inherit several structural limitations: reliance on specialized hardware or pre-scanned environments, heavy multi-stage processing, sensitivity to reconstruction failures and scale ambiguity, and difficulty scaling to unconstrained in-the-wild video.
The second family, exemplified by recent 3D-aware MLLMs, targets \emph{visual-based spatial intelligence} from monocular or multi-view RGB \cite{wu2025spatialmllm, fan2025vlm3r}.
They typically combine a general-purpose visual encoder \cite{bai2023qwenvl,zhai2023siglip} with a feed-forward geometry encoder \cite{wang2025cut3r,wang2025vggt} and then fuse the resulting features via simple mechanisms such as MLP projection or one-stage cross-attention.
These designs avoid explicit 3D supervision and are more scalable, but the way they integrate spatial cues remains largely ad hoc.

A closer look at these two lines of work reveals a common limitation.
Explicit-3D methods over-commit to precomputed geometry and do not address how viewpoint should interact with language-native reasoning.
Visual-based methods, on the other hand, tend to treat all geometric signals—image features, spatial features, and camera information—as if they belonged to a single homogeneous feature space.
In practice, camera-related signals are often appended as auxiliary embeddings or implicitly blended into fusion layers.
This conflation ignores a principle long emphasized in 3D vision \cite{sajjadi2022srt,zhang2025flare,wang2025vggt,li2025vicasplat,jiang2025anysplat}: \emph{camera} (ego/viewpoint) and \emph{scene} (allocentric content) play fundamentally different roles.

Motivated by this insight, we present \textbf{SpaceMind}, a large vision-language model explicitly designed for 3D spatial reasoning from RGB inputs.
SpaceMind treats the camera representation as a \emph{dedicated modality} that guides how spatial information is injected into the visual stream, rather than as a passive conditioning vector.
Concretely, SpaceMind uses a visual encoder \cite{chen2024internvl} and a spatial encoder \cite{wang2025vggt} to obtain semantic and geometry-aware tokens, and introduces a \emph{Camera-Guided Modality Fusion} (CGMF) module before the language model.
CGMF applies camera-conditioned biasing over spatial tokens, a query-independent spatial importance weighting, and camera-conditioned gating on fused features, making the role of the camera explicit in the fusion process while preserving the interface of standard VLMs.
This design is intentionally simple—built from standard components—but aligns viewpoint, spatial cues, and semantics within a unified multimodal framework.
Empirically, SpaceMind achieves a new state of the art on \textbf{VSI-Bench} \cite{yang2025vsi}, with a clear margin over existing open and proprietary systems, and delivers comparable or superior performance on other spatial reasoning benchmarks such as SQA3D \cite{ma2023sqa3d} and SPBench \cite{stogiannidis2025mind}, indicating strong generalization across datasets and task formats.
In summary, our contributions are four-fold:

\begin{itemize}
    \item We propose \textbf{SpaceMind}, a multimodal large language model tailored for 3D spatial reasoning from visual observations.
    \item We identify a conceptual gap in prior work: existing fusion paradigms conflate camera and scene features, whereas explicitly treating the camera representation as a guiding modality leads to more coherent spatial reasoning.
    \item We introduce a simple yet effective Camera-Guided Modality Fusion (CGMF) module that integrates camera, spatial, and visual tokens within a unified framework.
    \item We show that SpaceMind achieves new \textbf{state-of-the-art} performance across diverse spatial reasoning datasets and tasks, consistently outperforming all prior methods.

\end{itemize}  

\section{Related Work}

\subsection{MLLMs for Image and Video Understanding}

Multimodal large language models extend large language models to vision and other modalities by aligning visual and textual tokens in a shared embedding space.
CLIP \cite{radford2021clip} and ALIGN \cite{alayrac2022flamingo} learn joint image-text representations via contrastive pretraining, enabling strong zero-shot transfer.
Flamingo \cite{alayrac2022flamingo} and BLIP-2 \cite{li2023blip2} introduce token-level or feature-level fusion modules (e.g., Q-Former, cross-attention bridges) to couple frozen LLM backbones with vision encoders, 
while instruction-tuned systems such as the LLaVA family \cite{liu2023llava,liu2024llava,li2024llavaonevision,li2024llavainterleave,zhang2024llavanextvideo},
MiniGPT-4 \cite{zhu2023minigpt}, Qwen-VL \cite{bai2023qwenvl}, and InternVL \cite{chen2024internvl} further enhance open-ended visual QA.
Recent video MLLMs \cite{maaz2023videochatgpt,zhang2023videollama,lin2023videollava,song2023moviechat,weng2024longvlm} extend these designs to temporal inputs through spatiotemporal pooling, causal attention over frame tokens, or long-context architectures, and achieve strong performance on video understanding benchmarks.

Despite their success, these models are mostly optimized for semantic and temporal understanding rather than geometric reasoning.
They typically treat videos as sequences of 2D observations, with little explicit modeling of camera motion, global scene layout, or cross-view consistency.
As a result, they perform well on recognition-style tasks but remain unreliable for metric judgments or layout reconstruction, motivating architectures with stronger inductive bias toward 3D spatial understanding.

\subsection{3D Visual-Based Spatial Intelligence}

A growing body of work aims to endow multimodal models with 3D awareness.
One line adopts explicit 3D inputs \cite{zhu2024pq3d,zhu2023vista,yuan2025scener1,ma2024leo,man2024sig3d,wang2025scenellm,deng20253dllava,huang2024chatscene,zhu2025llava3d,zheng2025video3dllm}:
models take point clouds, meshes, RGB-D scans, or voxel-style features as input and align them with language using Q-Formers, 3D detectors, or volumetric aggregation.
These approaches demonstrate strong 3D grounding when high-quality geometry is available, but rely on depth sensors or reconstruction pipelines,
making them sensitive to reconstruction errors and less scalable to unconstrained monocular video.

Alongside these sensor-dependent designs, a second line of work including SpaceR \cite{ouyang2025spacer} and VILASR \cite{wu2025vilasr} attempts to elicit 3D reasoning directly from existing VLMs \emph{without} introducing additional geometry encoders.
Operating in the same RGB-only regime as our setting, these methods keep the native visual backbone mostly frozen and instead rely on carefully engineered data and training strategies, but improvements on rigorous spatial benchmarks remain modest.

Complementary work such as VLM-3R \cite{fan2025vlm3r} and Spatial-MLLM \cite{wu2025spatialmllm} augments VLMs with geometry-aware encoders that infer spatial tokens and camera tokens from images or short videos, followed by lightweight fusion layers.
These approaches demonstrate clear improvements on spatial benchmarks \cite{yang2025vsi,ma2023sqa3d}, validating the value of injecting geometric priors.
Yet, their fusion mechanisms typically mix camera and scene tokens within a shared space via simple concatenation or single-stage cross-attention, without explicitly modeling their distinct roles.
Our work follows the same RGB-only, geometry-augmented paradigm but introduces a fusion design where the camera representation serves as an explicit guiding modality for spatial tokens.

\subsection{3D Reconstruction from Images}

Progress in 3D reconstruction \cite{zhang2025flare,wang2025pi3,mildenhall2020nerf,kerbl2023gaussiansplatting} provides the geometric priors that many 3D-aware multimodal systems exploit.
Classical pipelines based on Structure-from-Motion \cite{schonberger2016sfm} and Multi-View Stereo \cite{furukawa2010mvs} recover camera poses and dense depth through feature matching, bundle adjustment, and volumetric or patch-based aggregation, but require multi-stage optimization and often assume controlled capture conditions.
Learnable MVS approaches, such as MVSNet and its variants \cite{yao2018mvsnet,gu2020cascade}, predict per-view depth or cost volumes under known cameras, improving efficiency and accuracy while still depending on calibrated multi-view input.

Feed-forward visual-geometry models such as DUSt3R \cite{wang2024dust3r}, MASt3R \cite{leroy2024mast3r}, and VGGT \cite{wang2025vggt} directly infer pixel-wise 3D correspondences, point maps, and camera parameters from image pairs or short sequences, often without requiring explicit calibration.
By casting reconstruction as a single forward pass of a transformer-style backbone, these models relax assumptions on camera poses, scale to in-the-wild data, and unify tasks such as depth estimation, camera pose recovery, and dense point cloud reconstruction.
Such advances disentangle camera (viewpoint) and scene (geometry) representations and make it practical to obtain geometry-aware spatial tokens and camera tokens from ordinary RGB sequences, motivating architectures like SpaceMind that focus on fusing these signals to support reliable spatial reasoning in multimodal language models.

\begin{figure*}[!t]
    \centering
    \includegraphics[width=\textwidth]{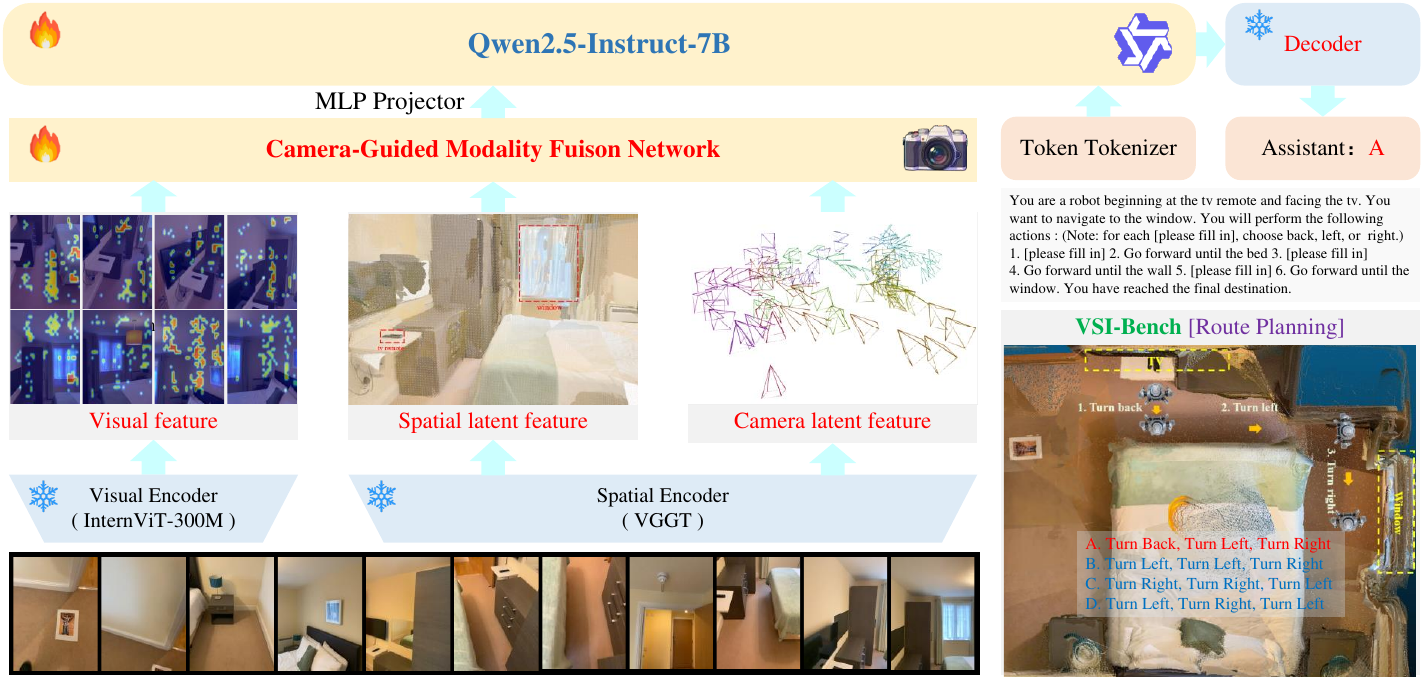}
    \caption{
    Overall pipeline of \textbf{SpaceMind}.
    Given a text prompt and an image sequence, a visual encoder produces semantic visual tokens, while a spatial encoder produces geometry-aware tokens together with per-frame camera tokens that summarize viewpoint information.
    The proposed Camera-Guided Modality Fusion (CGMF) module takes these three streams as input: it uses camera tokens to modulate spatial tokens, estimates their relative importance, and injects the resulting spatial cues into the visual tokens.
    The fused, view-aware visual tokens preserve the original token shape expected by the multimodal LLM, enabling SpaceMind to be trained end-to-end on RGB-only data while remaining compatible with standard VLM architectures and enhancing their 3D spatial reasoning ability.
    }
    \label{fig:pipeline}
\end{figure*}

\section{Method}

We propose SpaceMind, a multimodal large language model designed for spatial reasoning from visual observations.
Figure~\ref{fig:pipeline} shows the overall architecture of SpaceMind.
Given a text prompt $T$ and a sequence of images $S = \{I_i\}_{i=1}^N$, where each image $I_i \in \mathbb{R}^{H \times W \times 3}$ represents a view of the scene, the model performs reasoning grounded in the 3D environment.
We describe the overall architecture of SpaceMind in Sec.~\ref{sec:overview}, and detail its core component---the Camera-Guided Modality Fusion (CGMF) module---in Sec.~\ref{sec:cgmf}.

\subsection{SpaceMind Architecture}\label{sec:overview}
Our model \textbf{SpaceMind}, couples a strong vision--language backbone \cite{zhu2025internvl3,yang2024qwen2}
 with a feed-forward spatial encoder \cite{wang2025vggt}
and inserts a \emph{Camera-Guided Feature Modality Fusion} (CGMF) layer before the LLM.

In symbols, the input to our model consists of a text prompt $T$ (e.g., ``How many tables are in the scene?'') and a sequence of images $S = \{I_i\}_{i=1}^N$, where each $I_i \in \mathbb{R}^{H \times W \times 3}$ is a RGB image.

Let $g$ denote the LLM backbone, $e_v$ the visual encoder (which is the ViT of our backbone \cite{zhu2025internvl3}), and $e_s$ the spatial encoder (based on VGGT~\cite{wang2025vggt}).
SpaceMind performs the following computations:
\begin{align}
f_v &= e_v(\{I_i\}_{i=1}^N), \\
f_s, f_c, f_{\text{register}} &= e_s(\{I_i\}_{i=1}^N),
\end{align}
where $p_v$ and $p_s$ denote the patch sizes of $e_v$ and $e_s$, respectively, and $M_v = \left\lfloor \frac{H}{p_v} \right\rfloor \times \left\lfloor \frac{W}{p_v} \right\rfloor$, $M_s = \left\lfloor \frac{H}{p_s} \right\rfloor \times \left\lfloor \frac{W}{p_s} \right\rfloor$.
Thus, $f_v \in \mathbb{R}^{N \times M_v \times d_v}$, $f_s \in \mathbb{R}^{N \times M_s \times d_s}$, $f_c \in \mathbb{R}^{N \times 1 \times d_s}$, and $f_{\text{register}} \in \mathbb{R}^{N \times 4 \times d_s}$.

Following the setup in VGGT~\cite{wang2025vggt}, we discard $f_{\text{register}}$ and perform the following feature fusion:
\begin{equation}
f_{\text{fused}} = F(f_v, f_s, f_c), \quad f_{\text{fused}} \in \mathbb{R}^{N \times M_v \times d_v},
\end{equation}
where $F$ denotes the CGMF module (detailed in Sec.~\ref{sec:cgmf}).

The response of our model is then obtained by:
\begin{equation}
R = g(f_{\text{fused}}, T).
\end{equation}

\noindent
\textbf{Design principle.} Instead of concatenating camera cues with spatial features as in prior work~\cite{fan2025vlm3r,wu2025spatialmllm},
we expose camera tokens as a separate control modality that guides how spatial information influences visual tokens,
introducing a lightweight inductive bias from viewpoint to geometry without hand-crafted constraints.

\subsection{Camera-Guided Modality Fusion (CGMF)}\label{sec:cgmf}

The CGMF module $F$ fuses spatial features $f_s$ with visual features $f_v$ under the guidance of the camera latent $f_c$.
Formally, we take
\[
f_v \in \mathbb{R}^{N \times M_v \times d_v},\quad
f_s \in \mathbb{R}^{N \times M_s \times d_s},\quad
f_c \in \mathbb{R}^{N \times 1 \times d_s},
\]
and produce $f_{\text{fused}} \in \mathbb{R}^{N \times M_v \times d_v}$ to maintain the same input dimensionality as $f_v$, minimizing the impact on the pre-trained LLM's learned feature distribution.

We first project all inputs into a shared attention space with width $d_a$:
\begin{gather}
Q = P_Q(\mathrm{LN}(f_v)), \\
K = P_K(\mathrm{LN}(f_s)), \quad
V = P_V(\mathrm{LN}(f_s)), \\
C = P_C(f_c),
\end{gather}
where $Q \in \mathbb{R}^{N \times M_v \times d_a}$,
$K,V \in \mathbb{R}^{N \times M_s \times d_a}$,
and $C \in \mathbb{R}^{N \times 1 \times d_a}$.

Then, we apply camera-conditioned spatial bias.
A core observation from recent 3D reconstruction models \cite{wang2024dust3r,leroy2024mast3r,wang2025vggt,zhang2025flare} is that the disentanglement of camera and scene features is usually beneficial to both types of information.
To inject this spirit in a simple and learnable way, we use camera tokens to modulate spatial tokens.
For each spatial token, we concatenate its feature with the corresponding frame-wise camera token and apply a small MLP:
\begin{gather}
B_g = \mathrm{MLP}\big([f_s,\ f_c]\big), \\
B_g \in \mathbb{R}^{N \times M_s \times d_a},
\end{gather}
and update
\begin{gather}
K \leftarrow K + B_g,\quad
V \leftarrow V + B_g.
\end{gather}
This enables camera-aware adjustment of spatial keys and values, letting the model encode viewpoint-dependent structure directly into attention keys and values.

\begin{figure*}[!t]
    \centering
    \includegraphics[width=\textwidth]{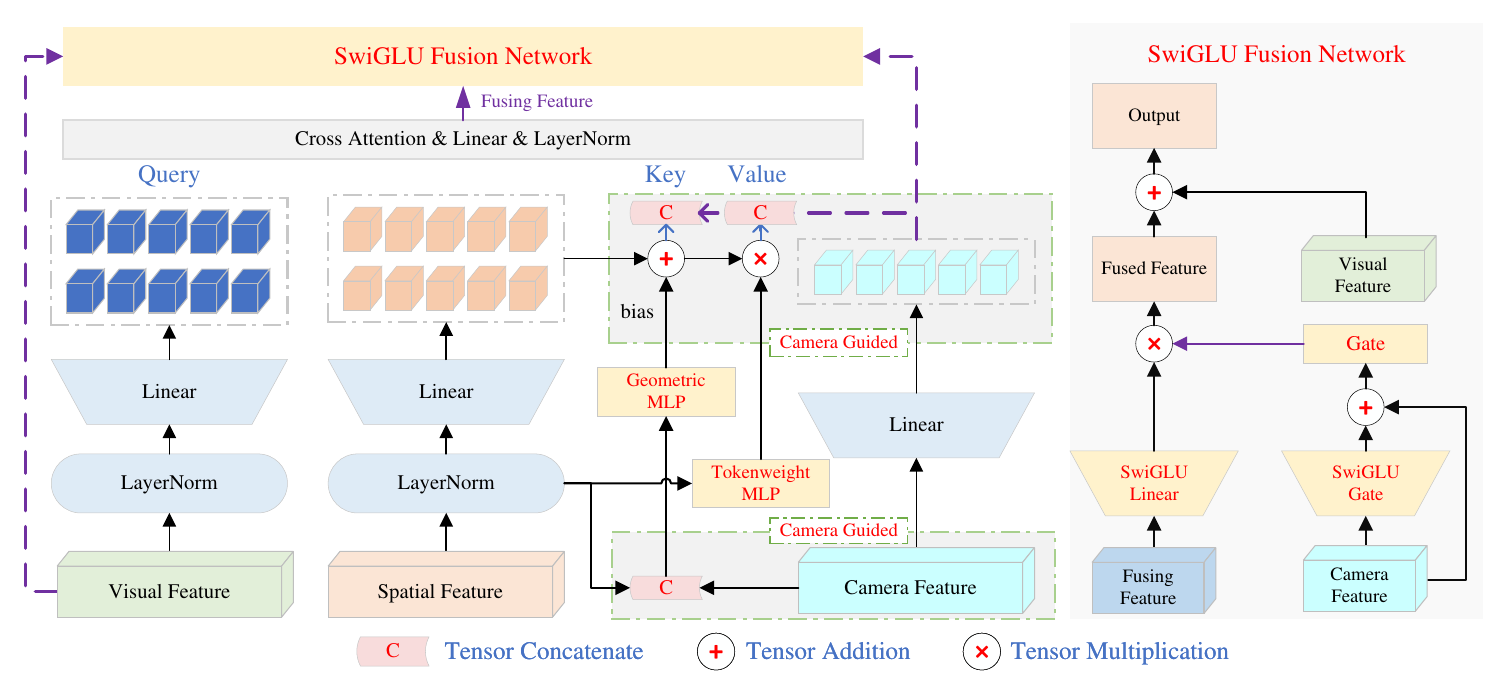}
    \caption{The architecture of the CGMF module. CGMF takes visual tokens $f_v$, spatial tokens $f_s$, and camera tokens $f_c$ as input, and outputs fused visual tokens with the same shape as $f_v$. The camera features are leveraged to refine the construction of geometric residuals, which further guide the cross-attention fusion between visual and spatial features. In addition, the SwiGLU Fusion Network follows the SwiGLU mechanism to achieve efficient multimodal feature fusion under the guidance of camera information.}
    \label{fig:cgmf}
\end{figure*}

Next, we compute per-token spatial importance.
Recent 3D reconstruction models such as DUSt3R \cite{wang2024dust3r} and VGGT \cite{wang2025vggt} output per-pixel confidence estimates that indicate the reliability of geometric predictions.
Motivated by this, we predict a per-token importance weight to mimic this behavior at the token level:
\begin{gather}
W_t = \sigma(\mathrm{MLP}(f_s)), \\
W_t \in \mathbb{R}^{N \times M_s \times 1},
\end{gather}
and rescale
\begin{gather}
V \leftarrow V \odot W_t.
\end{gather}
This term can be viewed as a soft mixture of reliability and importance: it reflects how useful each spatial token is as geometric evidence before seeing the query.
Unlike standard attention scores, which depend jointly on queries and keys, $W_t$ is determined purely from the spatial branch and thus provides a query-independent prior over spatial tokens for the subsequent fusion, where visual queries $Q$ attend over the camera-modulated and importance-weighted spatial keys and values.

We then insert the camera tokens into the attention memory:
\begin{gather}
K' = [C; K], \quad
V' = [C; V],
\end{gather}
where $[\cdot;\cdot]$ denotes concatenation along the token dimension.
Visual queries $Q$ attend over the concatenated memory:
\begin{gather}
\hat{f} = \mathrm{Attn}(Q, K', V'),
\end{gather}
yielding $\hat{f} \in \mathbb{R}^{N \times M_v \times d_a}$ that combines semantic, spatial, and camera information.

Finally, we map back to the visual feature space and apply a camera-conditioned gate.
Inspired by the SwiGLU \cite{shazeer2020glu,dauphin2017glu} activation used in modern LLMs \cite{chowdhery2022palm,hoffmann2022chinchilla,touvron2023llama,touvron2023llama2,yang2024qwen2,groeneveld2024olmo},
we split the camera projection into a Swish branch and a linear branch:
\begin{gather}
f_{\text{proj}} = \mathrm{LN}(P_O(\hat{f})),\\
\bar{C} = \mathrm{squeeze}(C) \in \mathbb{R}^{N \times d_a},\\
u = P_{g,1}(\bar{C}) \in \mathbb{R}^{N \times d_v},\\
v = P_{g,2}(\bar{C}) \in \mathbb{R}^{N \times d_v},\\
g = \mathrm{Swish}(u) \odot v,
\end{gather}
where $\mathrm{Swish}(x) = x \cdot \sigma(x)$, and the multiplicative form
$\mathrm{Swish}(u) \odot v$ follows the standard SwiGLU gating design.

The camera-conditioned gate $g$ is then broadcast over visual tokens and used to modulate the fused features:
\begin{gather}
f_{\text{fused}} = P_L(f_{\text{proj}}) \odot g[:,\text{None},:] + f_v.
\end{gather}
Here the camera embedding serves as the conditioning signal that controls how strongly spatially enriched features influence the final representation.


\begin{table*}[!t]
    \centering
    \small
    \caption{\textbf{Evaluation on VSI-Bench \cite{yang2025vsi}.}
    SpaceMind sets a new state-of-the-art, achieving the best average score and outperforming all prior models on every individual subtask, often by a large margin.
    }
    \setlength{\tabcolsep}{4pt}
    \resizebox{\textwidth}{!}{
    \begin{tabular}{lccccccccc}
        \toprule
        Methods & Avg. & \multicolumn{4}{c}{Numerical Question} & \multicolumn{4}{c}{Multiple-Choice Question} \\
        & & Obj. Cnt. & Abs. Dist. & Obj. Size & Room Size & Rel. Dist. & Rel. Dir. & Route Plan & Appr. Order \\
        \midrule
        \multicolumn{10}{l}{\emph{Proprietary Models (API)}} \\
        GPT-4o \cite{hurst2024gpt4o} & 34.0 & 46.2 & 5.3 & 43.8 & 38.2 & 37.0 & 41.3 & 31.5 & 28.5 \\
        Gemini-1.5 Flash \cite{gemini2024gemini15} & 42.1 & 49.8 & 30.8 & 53.5 & 54.4 & 37.7 & 41.0 & 31.5 & 37.8 \\
        Gemini-1.5 Pro \cite{gemini2024gemini15} & 45.4 & 56.2 & 30.9 & 64.1 & 43.6 & 51.3 & 46.3 & 36.0 & 34.6 \\
        \midrule
        \multicolumn{10}{l}{\emph{Open-source VLMs}} \\
        InternVL3-78B \cite{zhu2025internvl3} & \textbf{48.5} & \textbf{71.2} & \textbf{53.7} & 44.4 & \textbf{39.5} & \textbf{55.9} & 39.5 & 28.9 & \textbf{54.5} \\
        LLaVA-NeXT-Video-7B \cite{zhang2024llavanextvideo} & 35.6 & 48.5 & 14.0 & 47.8 & 24.2 & \underline{43.5} & \textbf{42.4} & \underline{34.0} & 30.6 \\
        LLaVA-NeXT-Video-72B \cite{zhang2024llavanextvideo}    & \underline{40.9} & \underline{48.9} & 22.8 & \underline{57.4} & 35.3 & 42.4 & 36.7 & \textbf{35.0} & \underline{48.6} \\
        QWen2.5VL-7B \cite{bai2025qwen25vl}            & 33.0 & 40.9 & 14.8 & 43.4 & 10.7 & 38.6 & 38.5 & 33.0 & 29.8 \\
        LLaVA-OneVision-7B \cite{li2024llavaonevision}      & 32.4 & 47.7 & 20.2 & 47.4 & 12.3 & 42.5 & 35.2 & 29.4 & 24.4 \\
        LLaVA-OneVision-72B \cite{li2024llavaonevision}     & 40.2 & 43.5 & \underline{23.9} & \textbf{57.6} & \underline{37.5} & 42.5 & \underline{39.9} & 32.5 & 44.6 \\
        \midrule
        \multicolumn{10}{l}{\emph{Specialized Spatial Reasoning Models}} \\
        Spacer \cite{ouyang2025spacer}                  & 45.5 & 57.8 & 28.2 & 59.9 & 47.1 & 40.1 & 45.4 & 33.5 & \underline{52.1} \\
        ViLaSR \cite{wu2025vilasr}                  & 45.4 & 63.5 & 34.4 & 60.6 & 30.9 & 48.9 & 45.2 & 30.4 & 49.2 \\
        Spatial-MLLM \cite{wu2025spatialmllm}          & 48.4 & 65.3 & 34.8 & 63.1 & 45.1 & 41.3 & 46.2 & 33.5 & 46.3 \\
        VLM-3R \cite{fan2025vlm3r}               & \underline{60.9} & \underline{70.2} & \underline{49.4} & \underline{69.2} & \underline{67.1} & \underline{65.4} & \underline{80.5} & \textbf{45.4} & 40.1 \\
        \textbf{SpaceMind (Ours)}  & \textbf{69.6} & \textbf{73.3} & \textbf{61.4} & \textbf{77.3} & \textbf{74.2} & \textbf{67.2} & \textbf{88.4} & \underline{44.3} & \textbf{70.6} \\
        \bottomrule
    \end{tabular}
    }
    \label{tab:vsibench}
\end{table*}

In summary, CGMF realizes a camera-guided fusion pattern using three simple ingredients:
a camera-conditioned bias on spatial tokens, a query-independent weighting over spatial patches, and a camera-conditioned gating of the fused representation.
All operations are lightweight and compatible with existing VLM implementations, yet together they encourage the model to align spatial reasoning with how the scene is viewed.

\section{Experiments}

\subsection{Implementation Details}\label{sec:implementation}

\noindent
\textbf{Backbone.}
SpaceMind adopts InternVL3-8B \cite{zhu2025internvl3} as the language model backbone and its paired image encoder, 
InternViT-300M \cite{zhu2025internvl3}, as the visual encoder.
For geometric understanding, we use the aggregator module of VGGT \cite{wang2025vggt} as the spatial encoder.

\noindent
\textbf{Training setup.}
The model is fine-tuned for two epochs on a mixture of datasets, including VLM-3R-data \cite{fan2025vlm3r}, ViCA-322K \cite{feng2025vica}, and the training split of SQA3D \cite{ma2023sqa3d}.
During fine-tuning, we freeze both the visual and spatial encoders, fully update the proposed CGMF module, and apply Low-Rank Adaptation (LoRA) \cite{hu2022lora} with rank $256$ and scaling factor $512$ to the InternVL3-8B language model backbone.
We employ a cosine learning-rate scheduler with an initial learning rate of $2\times10^{-5}$ and a warm-up ratio of $0.03$, and train with a global batch size of $64$.
The full training process required roughly $25$ hours on a cluster of $64$ NVIDIA H100 (80GB) GPUs.

\noindent
\textbf{Data preprocessing.}
For both training and inference, SpaceMind uniformly samples $34$ frames from each scene, discards the first and last frames, and uses the remaining $32$ frames as input.
Each image is resized to $448\times448$ for InternViT and zero-padded to $518\times518$ for VGGT, following its original configuration.

\subsection{Evaluations}

We evaluate SpaceMind on three benchmarks, including VSI-Bench \cite{yang2025vsi}, SQA3D \cite{ma2023sqa3d} (in-domain), 
and SPBench \cite{li2025spbench} (out-of-domain).
We will release code and model checkpoints to support future research.

\begin{table*}[!t]
    \centering
    \small
    \caption{\textbf{Evaluation on SQA3D \cite{ma2023sqa3d} test split.} 
    SpaceMind achieves the best performance across most question types and establishes a new state of the art on both EM@1 and EM@R1, despite using video-only inputs while many existing methods rely on richer modalities.
    }
    \setlength{\tabcolsep}{9pt}
    \begin{tabular}{lcccccccc|c}
        \toprule
        Method & \multicolumn{6}{c}{Test set} & EM@1 & EM@R1 & \textbf{Video-Input Only} \\
        \cmidrule(lr){2-7}
        & What & Is & How & Can & Which & Others & & & \\
        \midrule
        PQ3D \cite{zhu2024pq3d}      & 37.1 & 61.3 & 44.5 & 60.9 & 47.0 & 45.1 & 47.1 & 49.3   & No \\
        3D-VisTA \cite{zhu2023vista}      & 34.8 & 63.3 & 45.4 & 69.8 & 47.2 & 48.1 & 48.5 & 50.9   & No \\
        LEO \cite{ma2024leo}           & 39.0   & 63.9   & 44.9   & 66.2   & 47.7   & 51.1   & 50.0 & 52.4 & No \\
        SIG3D \cite{man2024sig3d}           & 35.6   & 67.2   & 48.5   & 71.4   & 49.1   & 45.8   & 52.6 & 54.4 & No \\
        Scene-LLM \cite{wang2025scenellm}     & 40.9 & 69.1 & 45.0 & 70.8 & 47.2 & 52.3 & 54.2 & 56.2   & No \\
        ChatScene \cite{huang2024chatscene}     & 45.4 & 67.0 & 52.0 & 69.5 & 49.9 & \underline{55.0} & 54.6 & 57.5 & No \\
        Video-3D LLM \cite{zheng2025video3dllm}  & \underline{51.1} & \underline{72.4} & \underline{55.5} & \underline{69.8}
                      & \underline{51.3} & \textbf{56.0} & \underline{58.6} & \underline{60.8}   & No \\
        \textbf{SpaceMind (Ours)}
                      & \textbf{54.1} & \textbf{74.8} & \textbf{61.7} & \textbf{71.0}
                      & \textbf{51.9} & 53.6 & \textbf{61.1} & \textbf{63.8} & Yes \\
        \bottomrule
    \end{tabular}%
    \label{tab:sqa3d}
\end{table*}

\begin{table*}[!t]
    \centering
    \small
    \caption{\textbf{Evaluation on SPBench \cite{li2025spbench}.}
    All models are evaluated without using SPBench training data. SpaceMind achieves the best performance on both SPBench-SI and SPBench-MV, outperforming general-purpose VLMs and prior spatial models by a clear margin.
    }
    \vspace{-5pt}
    \setlength{\tabcolsep}{16pt}
    \begin{tabular}{lccccccc}
        \toprule
        Methods & \multicolumn{3}{c}{SPBench-SI} & \multicolumn{3}{c}{SPBench-MV} & Overall \\
        \cmidrule(lr){2-4} \cmidrule(lr){5-7}
         & NQ & MCQ & Avg. & NQ & MCQ & Avg. & \\
        \midrule
        \multicolumn{8}{l}{\emph{Proprietary Models}} \\
        GPT-4o \cite{hurst2024gpt4o} & 24.5 & 60.3 & 42.4 & 40.7 & 59.4 & 50.1 & 46.2 \\
        Gemini-2.0-Flash \cite{gemini2024gemini20} & 49.0 & 60.4 & 54.7 & 51.9 & 50.7 & 51.3 & 53.0 \\
        \midrule
        \multicolumn{8}{l}{\emph{Open-Source Models}} \\
        InternVL-2.5-4B \cite{chen2024internvl25} & \underline{31.8} & 53.3 & \underline{42.6} & \textbf{37.7} & 51.4 & \textbf{44.6} & \underline{43.6} \\
        InternVL-2.5-8B \cite{chen2024internvl25} & 28.3 & \underline{56.3} & 42.3 & \underline{37.3} & 47.5 & \underline{42.4} & 42.3 \\
        Kimi-VL-A3B \cite{du2025kimivl} & 25.7 & 44.9 & 35.3 & 23.3 & \textbf{57.6} & 40.5 & 37.9 \\
        LLaVA-OneVision-7B \cite{li2024llavaonevision} & 25.4 & 41.0 & 33.2 & 20.6 & 49.6 & 35.1 & 34.2 \\
        Qwen2.5-VL-7B \cite{bai2025qwen25vl} & \textbf{36.3} & \textbf{60.5} & \textbf{48.4} & 28.9 & 49.8 & 39.3 & \textbf{43.9} \\
        Qwen2.5-VL-3B \cite{bai2025qwen25vl} & 24.3 & 56.2 & 40.3 & 25.6 & \underline{53.2} & 39.4 & 39.8 \\
        \midrule
        \multicolumn{8}{l}{\emph{Spatial Reasoning Models}} \\
        Video-R1 \cite{feng2025videor1} & 27.7 & \underline{62.0} & 44.9 & 32.5 & 53.0 & 42.8 & 43.8 \\
        SpaceR-7B \cite{ouyang2025spacer} & 35.7 & 61.5 & 48.6 & 63.2 & 53.7 & 58.5 & 53.5 \\
        VILASR-7B \cite{wu2025vilasr} & 36.6 & \textbf{63.7} & \underline{50.2} & 56.2 & \underline{59.6} & 57.9 & \underline{54.0} \\
        Spatial-MLLM-4B \cite{wu2025spatialmllm} & \underline{38.1} & 49.3 & 43.7 & \underline{63.7} & 58.9 & \underline{61.3} & 52.5 \\
        \textbf{SpaceMind (Ours)} & \textbf{66.3} & 53.2 & \textbf{59.7} & \textbf{76.2} & \textbf{70.5} & \textbf{73.8} & \textbf{67.3} \\
        \bottomrule
    \end{tabular}
    \label{tab:spbench}
    \vspace{-8pt}
\end{table*}

\noindent
\textbf{Evaluation on VSI-Bench.}
VSI-Bench \cite{yang2025vsi} contains over $5{,}000$ questions curated from ARKitScenes \cite{baruch2021arkitscenes}, ScanNet \cite{dai2017scannet}, and ScanNet++ \cite{yeshwanth2023scannetpp}.
It comprises two answer formats—Numerical Answer (NA) and Multiple-Choice Answer (MCA)—covering eight subtasks: object counting, relative direction, absolute distance, route planning, object size, room size, relative distance, and appearance order.
Following the official protocol, we evaluate NA with relative accuracy and MCA with mean accuracy.
As shown in Table~\ref{tab:vsibench}, \textbf{SpaceMind} attains the best score on nearly all eight subtasks and raises the overall average to \textbf{69.6}, a \textbf{+8.7} improvement over the strongest prior baseline (VLM-3R-7B, 60.9).
Gains appear in both NA and MCA groups: measurement-oriented tasks (absolute/relative distance, object/room size) improve consistently, while view-integration tasks show especially large gains on \emph{relative direction} and \emph{appearance order}.
Data-centric RGB-only baselines such as SpaceR \cite{ouyang2025spacer} and VILASR \cite{wu2025vilasr}, which keep the visual backbone largely fixed and focus on curriculum design, achieve averages of only $45.5$ and $45.4$ respectively—barely matching a generic VLM and still performing poorly on geometry-heavy subtasks like absolute/relative distance and appearance order—indicating that training and data tricks alone are insufficient to elicit robust 3D reasoning from 2D features.
Geometry-augmented models such as Spatial-MLLM \cite{wu2025spatialmllm} and VLM-3R \cite{fan2025vlm3r} narrow the gap by introducing explicit spatial tokens, but their one-stage fusion of camera and scene representations still leaves substantial headroom: they remain notably weaker than SpaceMind on view-integration metrics such as \emph{relative direction} and especially \emph{appearance order}.
In particular, \emph{appearance order} improves by \textbf{+30.5} points—the largest increase among all subtasks—suggesting that explicitly conditioning spatial tokens on the camera signal helps consolidate cross-view evidence and stabilize ordering judgments across viewpoints, while \emph{route planning} remains highly competitive with the prior state-of-the-art.

\noindent
\textbf{Evaluation on SQA3D.}
SQA3D \cite{ma2023sqa3d} is a situated 3D question-answering benchmark built from reconstructed indoor scenes in ScanNet \cite{dai2017scannet}.
It provides textual situations and questions that require localizing an egocentric viewpoint and reasoning about nearby objects, relations, and navigation targets.
The test split covers diverse question templates (\emph{What}, \emph{Is}, \emph{How}, \emph{Can}, \emph{Which}, and \emph{Others}).
We follow the standard protocol and report exact-match accuracy (EM@1) and its refined variant (EM@R1).
As shown in Table~\ref{tab:sqa3d}, SpaceMind achieves the best performance across most question types and sets a new state of the art on both metrics.
Unlike prior methods that rely on depth, point clouds, meshes, or other auxiliary modalities, SpaceMind uses only video inputs.
This demonstrates that our camera-guided fusion can recover strong 3D spatial cues directly from RGB videos, enabling robust situated reasoning under realistic video-only settings where dense 3D supervision is unavailable.

\begin{table*}[t!]
    \centering
    \small
    \caption{\textbf{Ablation Study on VSI-Bench.} 
    We analyze the contribution of each component in SpaceMind: (1) adding VGGT spatial tokens via a \emph{shallow cross-attention} fusion layer, (2) incorporating the \emph{token weight MLP} (twMLP), and (3) further introducing the \emph{geometric MLP} (geoMLP).
    Performance improves consistently as each module is added, and the full SpaceMind architecture achieves the highest accuracy not only on average, but also across all numerical and most multiple-choice subtasks, demonstrating the effectiveness of our model design.
    }
    \setlength{\tabcolsep}{4pt}
    \resizebox{\textwidth}{!}{
    \begin{tabular}{lccccccccc}
        \toprule
        Methods & Avg. & \multicolumn{4}{c}{Numerical Question} & \multicolumn{4}{c}{Multiple-Choice Question} \\
        & & Obj. Cnt. & Abs. Dist. & Obj. Size & Room Size & Rel. Dist. & Rel. Dir. & Route Plan & Appr. Order \\
        \midrule
        InternVL3-8B ft         & 63.07 & 70.41 & 50.95 & 74.42 & 64.82 & 64.42 & 72.90 & 40.19 & 66.47 \\
        InternVL3-8B + VGGT         & 66.77 & 71.83 & 58.98 & 74.51 & 70.07 & 65.30 & 85.17 & 41.89 & 66.42 \\
        InternVL3-8B + VGGT + twMLP & 67.17 & 72.68 & 58.14 & 75.11 & 72.12 & 65.85 & 84.96 & 41.46 & 67.05 \\
        InternVL3-8B + VGGT + twMLP + geoMLP & \underline{68.73} & \underline{72.90} & \underline{60.52} & \underline{76.39} & \underline{73.78} & \textbf{67.37} & \underline{87.64} & \underline{43.08} & \underline{68.20} \\
        \textbf{SpaceMind (full)}  & \textbf{69.58} & \textbf{73.31} & \textbf{61.37} & \textbf{77.35} & \textbf{74.20} & \underline{67.18} & \textbf{88.38} & \textbf{44.33} & \textbf{70.55} \\

        \bottomrule
    \end{tabular}
    }
    \label{tab:ablation}
\end{table*}

\noindent
\textbf{Evaluation on SPBench.}
We assess out-of-domain generalization on SPBench \cite{li2025spbench}, which is derived from the ScanNet \cite{dai2017scannet} validation split via the SpatialLadder-26k pipeline and contains two subsets: SPBench-SI (single-view) and SPBench-MV (multi-view).
Each subset includes numerical questions (NQ) and multiple-choice questions (MCQ) spanning absolute/relative distance, relative direction, object size, and (for SPBench-MV) cross-view object counting.
Following the official protocol, we report mean relative accuracy for NQ and accuracy for MCQ; the SI/MV scores are the arithmetic mean of their NQ and MCQ results, and the overall score is the average of SI and MV.
Importantly, SPBench is not included in our training data and thus serves as a genuine out-of-domain evaluation.
As shown in Table~\ref{tab:spbench}, SpaceMind substantially outperforms both general-purpose VLMs and prior specialized spatial models, establishing a new state of the art.
Notably, on the \emph{single-image} subset, SpaceMind outperforms all competing methods by a clear margin \emph{despite being trained exclusively on 32-frame clips per QA}, indicating strong cross-regime transfer and that our camera-guided fusion remains effective even when the input collapses to a single view.

\subsection{Ablation Studies}

We conduct ablation studies on VSI-Bench to understand how each component of SpaceMind contributes to 3D spatial reasoning.
All variants are trained under the same protocol described in Sec.~\ref{sec:implementation}, and the results are summarized in Table~\ref{tab:ablation}.

\noindent
\textbf{From RGB-only VLM to shallow spatial fusion.}
Directly fine-tuning InternVL3-8B \cite{zhu2025internvl3} achieves an average accuracy of $63.07$, indicating that a pure RGB-based VLM still struggles on geometry-heavy tasks.
Adding VGGT \cite{wang2025vggt} spatial tokens and fusing them with visual tokens through a shallow cross-attention layer (\textit{InternVL3-8B + VGGT}) boosts the average score to $66.77$ (+$3.70$).
The improvements are especially pronounced on numerically grounded subtasks such as \textit{Abs.\ Dist.} (from $50.95$ to $58.98$, +$8.03$) and \textit{Room Size} (from $64.82$ to $70.07$, +$5.25$), as well as \textit{Rel.\ Dir.} (from $72.90$ to $85.17$).
This confirms that dense geometric cues provide complementary depth and layout information beyond 2D ViT features.

\noindent
\textbf{Effect of token-weight MLP (twMLP).}
The token-weight MLP in CGMF predicts a query-independent reliability prior over spatial tokens (Sec.~\ref{sec:cgmf}).
Enabling this module (\textit{InternVL3-8B + VGGT + twMLP}) further improves the average accuracy to $67.17$.
Although the overall gain over the shallow fusion baseline is modest (+$0.40$), the variant with twMLP yields more consistent improvements across subtasks, especially on \textit{Room Size} (from $70.07$ to $72.12$) and several multiple-choice tasks.
This behavior matches our design intuition: geometric estimates can be noisy or unevenly distributed in cluttered indoor scenes, and a learned importance prior helps the model down-weight unreliable spatial regions before attention is computed.

\noindent
\textbf{Effect of geometric MLP (geoMLP).}
We then introduce the geometric MLP, which injects camera-conditioned bias into spatial keys and values by jointly processing spatial and camera tokens (Sec.~\ref{sec:cgmf}).
This configuration (\textit{InternVL3-8B + VGGT + twMLP + geoMLP}) raises the average accuracy to $68.73$, a clear improvement over the twMLP-only variant (+$1.96$ from $66.77$ and +$1.56$ from $67.17$).
The gains are concentrated on viewpoint-sensitive subtasks: \textit{Abs.\ Dist.} increases from $58.14$ to $60.52$, and \textit{Room Size} from $72.12$ to $73.78$.
These trends support our hypothesis that explicitly re-centering spatial tokens with respect to the current camera viewpoint yields more stable metric reasoning.

\noindent
\textbf{Full CGMF and camera-conditioned SwiGLU fusion.}
Finally, SpaceMind applies the camera-conditioned SwiGLU fusion network to gate the fused representation before feeding it into the LLM (Sec.~\ref{sec:cgmf}).
This full configuration (\textit{SpaceMind (full)}) achieves the best performance, with an average accuracy of $69.58$.
Compared to the geoMLP variant, we observe consistent improvements across almost all subtasks, including \textit{Abs.\ Dist.} (from $60.52$ to $61.37$), \textit{Rel.\ Dir.} (from $87.64$ to $88.38$), \textit{Route Plan} (from $43.08$ to $44.33$), and \textit{Appr.\ Order} (from $68.20$ to $70.55$).
These results indicate that the final camera-conditioned gate helps the model calibrate how strongly spatially enriched features should influence the visual backbone, and that the three ingredients of CGMF---spatial token weighting, camera-conditioned geometric bias, and camera-guided SwiGLU gating---work synergistically to enhance spatial reasoning.

\section{Conclusion}

We propose \textbf{SpaceMind}, a multimodal large language model for 3D spatial reasoning that revisits fusion from a camera-centric perspective.
Rather than collapsing geometry-related signals into a single stream, SpaceMind treats the camera representation as a guiding modality and integrates it via the proposed Camera-Guided Modality Fusion (CGMF) module.
CGMF applies camera-conditioned biasing, query-independent weighting over spatial evidence, and camera-conditioned gating while preserving the interface of standard VLMs and operating purely on RGB inputs.
Empirically, SpaceMind achieves a new state of the art on VSI-Bench with a clear margin and delivers comparable or superior results on other spatial reasoning benchmarks, indicating that how viewpoint and geometry are fused is a key factor in visuospatial intelligence.

{
    \small
    \bibliographystyle{ieeenat_fullname}
    \bibliography{main}
}

\end{document}